\documentclass[sigconf]{acmart}
\pdfoutput=1
\def\BibTeX{{\rm B\kern-.05em{\sc i\kern-.025em b}\kern-.08emT\kern-.1667em\lower.7ex\hbox{E}\kern-.125emX}}
\copyrightyear{2019}
\acmYear{2019}
\setcopyright{acmlicensed}
\acmConference[KDD '19]{KDD '98: ACM SIGKDD Conference On Knowledge Discovery and Data Mining}{August 04--08, 2019}{Anchorage, Alaska -- USA}
\acmBooktitle{KDD '19: ACM SIGKDD Conference On Knowledge Discovery and Data Mining, June 04--08, 2019, Anchorage, Alaska -- USA}
\acmPrice{15.00}
\acmDOI{10.1145/1122445.1122456}
\acmISBN{978-1-4503-9999-9/18/06}

\usepackage{amsmath,amsfonts,amssymb,amsthm}   
\usepackage{commath}
\usepackage{subcaption}
\usepackage{caption}
\usepackage{tikz}
\usetikzlibrary{chains, positioning, shapes.geometric, calc, arrows, arrows.meta}
\usepackage{graphicx}
\usepackage{todonotes}
\usepackage{multirow}
\usepackage{mathtools}

\newcommand{\eg}{\textit{e.g.}}
\newcommand{\cf}{\textit{cf.}}
\newcommand{\etc}{\textit{etc.}}

\newcommand{\pen}{K} 
\newcommand{\goal}{G} 

\newcommand{\market}{M} 

\newcommand{\cv}{a} 

\newcommand{\cost}{C} 
\newcommand{\loss}{\cost} 

\newcommand{\volume}{V} 
\newcommand{\spend}{S} 
\newcommand{\TotalVolume}[1]{\sum_{s=0}^{#1}{\volume_s}} 
\newcommand{\TotalSpend}[1]{\sum_{s=0}^{#1}{\spend_s}} 

\newcommand{\FS}{\hat{H}} 

\newcommand{\rsignal}{\hat{\volume}} 

\title{Recurrent Neural Networks for Stochastic Control\\in Real-Time Bidding}

\author{Nicolas Grislain}
\affiliation{\institution{Verizon Media}}
\email{nicolas.grislain@verizonmedia.com}

\author{Nicolas Perrin}
\affiliation{\institution{Verizon Media}}
\email{nicolas.perrin@verizonmedia.com}

\author{Antoine Thabault}
\affiliation{\institution{Verizon Media}}
\email{antoine.thabault@verizonmedia.com}

\date{\today}

\begin{abstract}
Bidding in real-time auctions can be a difficult stochastic control task;
especially if underdelivery incurs strong penalties and the market is very uncertain.

Most current works and implementations focus on optimally delivering a campaign given
a reasonable forecast of the market. Practical implementations have a feedback loop
to adjust and be robust to forecasting errors, but no implementation, to the best of our knowledge, uses a model of
market risk and actively anticipates market shifts.

Solving such stochastic control problems in practice is actually very challenging.
This paper proposes an approximate solution based on a \emph{Recurrent Neural Network} (RNN) architecture
that is both effective and practical for implementation in a production environment.
The RNN bidder provisions everything it needs to avoid missing its goal. It also
deliberately falls short of its goal when buying the missing impressions would cost more than
the penalty for not reaching it.
\end{abstract}



\begin{document}

\maketitle

\section{Introduction}
Since its advent in 2009, \emph{Real-Time Bidding} (RTB), \textit{a.k.a.} \emph{programmatic media buying}, has been growing very fast. In 2018, more than 80\% of digital display ads are bought programmatically in the US \cite{programmatic2018}.

RTB is allowing advertisers (buyers) to buy ad-spaces to digital publishers (sellers) at a very fine-grained level, down to the user and the particular ad impression. The problem of using all the information available about each user exposed to some ad-placement to deliver a certain amount of impressions, clicks or viewable impressions, in an optimal way, is called \emph{the optimal bidding problem}.

The optimal bidding problem may come in different flavors. It may be about maximizing a given \emph{Key Performance Indicator} (KPI): impressions, clicks or views, given a certain budget, or about minimizing the cost of reaching some KPI goal.
It is often formulated in a second price auction setup, but different setups, like first price auction or other exotic setups, are common on the market.

In this paper, we focus on the problem of optimizing one campaign, competing with the market in second price auctions.
The campaign is aiming at a daily KPI goal, with a penalty for falling short of the goal. This restriction does not harm the generality of this work
as most of it is generalizable to other sorts of goals and time spans.
The campaign is small enough so that the impact of its delivery on the market is negligible.

Market data is eventually observable, which means it is possible to know, after some time (possibly hours), at what price a given impression would have been bought even if it was lost.
This last assumption is valid in our practical setup where a company controls both an inventory and a bidder buying its own inventory on behalf of external advertisers, possibly in competition with third party bidders. This assumption can be relaxed though, at the cost of a more complex training process.

This paper puts a strong emphasis on the way market uncertainty is handled
in a context where a fixed goal is to be achieved despite the stochastic nature of the market.
Without uncertainty, our problem reduces to a relatively simple optimal control problem,
adding randomness makes it an intractable stochastic control problem.
In this paper we propose a characterization of the solution in terms
of a \emph{Partial Differential Equation} (PDE) and an approximate solution
using a \emph{Recurrent Neural Network} (RNN) representation.

The main contributions of this paper can be summarized as follows:
\begin{enumerate}
  \item It formalizes in section~\ref{section:bidding_problem} the optimal bidding problem as a stochastic control problem,
  where market volume and prices are stochastic,
  \item It solves numerically a simple case in section~\ref{section:toy_problem} and comments qualitatively the solutions,
  \item It builds a practical RNN that approximate the theoretical solution in section~\ref{section:mdp},
  \item The RNN is trained and tested at scale on a major ad-exchange
  as described in section~\ref{section:experiments}.
\end{enumerate}

\section{Related papers}

A description of the various challenges brought by the impression-level user-centric bidding compared to bulk, inventory-centric-buying is done in \cite{yuan2013real}.

\cite{zhang2014optimal, Zhang2016} gives a very broad overview of the optimal bidding problem.

\cite{chen2011real} solves a bidding problem with multiple campaigns and from the perspective of the publisher using linear programming and duality.
A similar question is solved in \cite{balseiro2014yield, jauvion2018optimal}. In those papers, the publisher wants to allocate impressions to campaigns in competition with third party RTB campaigns. \cite{jauvion2018optimal} allows for underdelivery by introducing a \emph{penalty} for underdelivery in its optimization program.

\cite{ghosh2009adaptive} describes a solution to the bidding problem with budget constraints and partially observed exchange.

To account for market uncertainty, the optimal bidding problem is solved using a \emph{Markov Decision Process} (MDP), constantly adapting to the new state of the campaign on the market.
\cite{gallego1994optimal} proposes a heuristic in the field of yield management.
\cite{Karlsson2014, Karlsson2016, Karlsson2018} propose to use a \emph{Proportional Integral} (PI) controller to control the bidding process and add some randomness to the bid to help exploration in a partially observed market and alleviate the exploration-exploitation dilemma.
\cite{cai2017real} uses dynamic programming to derive an optimal policy auction by auction.
Modelling the problem auction by auction, makes the proposed methodology slightly impractical.
\cite{fernandez2016optimal} gives a very rigorous statement of the problem and solves it in cases where
impressions are generated by homogeneous Poisson processes and market prices are \emph{independent and identically distributed} (IID).

The general bidding problem with nonstationary stochastic volume and partially observed market is a complex \emph{Reinforcement Learning} (RL) problem tackled in \cite{wu2018budget} using tools from the \emph{deep reinforcement learning} literature.
\cite{wu2018budget} uses, as is done in this paper, the common approach of bidding proportionally to the predicted KPI probability and solves a control problem over this proportionality factor every few minutes instead of optimizing for every impression.
It makes the approach practical for real uses.

\cite{wu2018budget} finds the use of immediate reward misleading during the training, pushing to solutions neglecting the budget constraint.
The approach proposed in this paper introduces budget constraints in the reward by simply adding a linear penalty. The bidder may explore the costly scenarios where it falls short of its goal and avoid them.

Also, the MDP trained in \cite{wu2018budget} uses a state engineered by the author, mainly: the current time step, the remaining budget, the budget consumption rate, the cost per mille at the last period, the last win-rate and the last reward. This choice is reasonable but the memory of the MDP is reduced to the remaining budget and what can be inferred from the last period.
The approach proposed in this paper does not specify the state space and state transition, the \emph{Recurrent Neural Network} (RNN) state is learned from the data. In particular \emph{it can learn and encode the type of day or the type of shocks the market is undergoing} and reacts accordingly.

\section{The bidding problem under uncertainty}
\label{section:bidding_problem}
In this section, the bidding problem is considered in the specific context of
a bidder aiming at delivering campaigns in competition with the market, on media owned by itself.
This does not harm the generality of the work, but explains the availability of sell-side data for
training and the kind of objective considered: number of impression at minimum cost.
Without sell-side data, the training of the model exposed below would be made more complex
by the censorship of market data for lost auctions.

\subsection{Formal statement of the bidding problem}

In this presentation, let $\left(\Omega, \mathcal{F}, \mathbb{P}\right)$ be a probability space
equipped with a filtration $\left(\mathcal{F}_t\right)_{t \in \mathbb{R}+}$.
A \emph{bidder} is assumed to bid against the market representing all the competition.
Let $\left(\mathcal{H}_t\right)_{t \in \mathbb{R}+}$ denote the sub-filtration encoding the restricted information accessible to the bidder.

All bidders receive ad requests modeled by the jumps of a Poisson process $\left(I_t\right)_t$ with intensity $\iota_t$.
For each impression opportunity $i$, happening at $t_i$ the bidder receives $x_i$\footnote{Variables $X$ are indifferently noted $X_{t_i}$ or $X_i$.}
the current features of the impression: \texttt{timestamp}, \texttt{auction\_id}, \texttt{user\_id},
\texttt{placement\_id}, \texttt{url}, \texttt{device}, \texttt{os}, \texttt{browser} or \texttt{geoloc}.

Based on $x_i$ and, more generally, on all past history $H_i \in \mathcal{H}_i$ a bid $a_i$ is chosen $a_i = a\left(H_i\right)$.
The bidder wins the auction whenever $a_i > b_i$, where $b_i$ is the highest of the other bids in the market.

Each impression has some value $u_i$ to the bidder. When trying to buy clicks, $u_i$ is $0$ or $1$ depending on the occurrence of a click.
The bidder is assumed to know its expected value $v_i(H_i) = \mathbb{E}\left(u_i|H_i\right)$.

In this paper, the bidder is assumed not to have a significant influence on the market.
The bidder has also access to some distribution of $b_i$ conditional on $\mathcal{H}_i$.

This paper characterizes an optimal bidding strategy $a\left(H_i\right)$.

\subsection{The bidding problem over a short period of time}

The bidder's spend follows the process:
\begin{equation*}
  \dif S_t = b_{I_t}\mathbf{1}_{a(H_t) > b_{I_t}} \dif I_t,
  \label{eq:inst_spend}
\end{equation*}
and the cumulative value follows:
\begin{equation*}
  \dif V_t = v(H_t)\mathbf{1}_{a(H_t) > b_{I_t}} \dif I_t.
  \label{eq:inst_value}
\end{equation*}

Let us consider a short period of time $\delta$ such that conditionally on $\mathcal{H}_t$,
$\iota_t$ is predictable with average value $\iota$ and $x_i$, $b_i$, $u_i$ are \emph{independent, identically distributed} (IID) over $[t, t+\delta]$.
Let us consider that $\delta$ and $\iota$ are such that $\delta \cdot \iota$ is large and its relative standard deviation small\footnote{
In practice $\delta$ would be in the order of magnitude of 100 \emph{seconds} while $\iota$ close to 1000 \emph{events per second} so the relative error would be around $1\%$.}:
\begin{equation*}
\frac{1}{\sqrt{\delta \cdot \iota}}<<1.
\end{equation*}

Over a period of time $[t, t+\delta]$, the set of impression is noted $\mathcal{I}_t$ and the number of impression is almost deterministic $I_t = \delta \cdot \iota_t$.
Because $x_i$, $b_i$, $u_i$ are IID, each auction is brand new and the only relevant information for the bidder is $x_i$, that is $a(H_i) = a(x_i)$.
In a Second Price Auction setup\footnote{See, \eg{}, \cite{roughgarden2016twenty} for an introduction to second price auctions.}, the spend is
\begin{equation*}
  S_{\mathcal{I}}\left(a\right) = \sum_{i \in \mathcal{I}} b_i \mathbf{1}_{a(x_i)>b_i}
  \label{eq:spend*}
\end{equation*}
and the value is
\begin{equation*}
  V_{\mathcal{I}}\left(a\right) = \sum_{i \in \mathcal{I}} u_i \mathbf{1}_{a(x_i)>b_i}.
  \label{eq:value}
\end{equation*}
Because the $x_i$, $b_i$ and $u_i$ are IID and the values summed over a large number of impressions, everything becomes deterministic and reduces to
\begin{equation*}
S_{\mathcal{I}}\left(a\right) \approx I \cdot \mathbb{E}_x\left[\int_0^{a(x)}b f\left(b; x\right) \dif b\right]
\end{equation*}
and
\begin{equation*}
V_{\mathcal{I}}\left(a\right) \approx I \cdot \mathbb{E}_x\left[v(x) F\left(a(x); x\right)\right],
\end{equation*}
where $f\left(b; x\right)$ is the \emph{Probability Density Function} (PDF) of $b$ conditional on $x$ and $F\left(b; x\right)$ is the \emph{Cumulative Density Function} (CDF) associated.

The optimization program of the bidder can be written
\begin{equation*}
\min_{a} C(a) = \min_{a} \left[ S_{\mathcal{I}}\left(a\right) + K\max\left(0, G-V_{\mathcal{I}}\left(a\right)\right) \right].
\end{equation*}
It can be read: \emph{the bidder chooses a bidding strategy $a$ such that its overall cost $\cost$
is minimized, while its goal $G$ (in number of impressions of clicks) is reached}.
The cost is composed of the spend $S_{\mathcal{I}}$ incurred by the purchase of impressions,
and a possible linear penalty $K\left(G-V_{\mathcal{I}}\left(a\right)\right)$ paid if one falls short
of the goal $G$\footnote{Note that the penalty does not have to be paid at the end of the short period of time, because the goal is additive, this short period of time can be combined with other periods of time as in the next section.}.
This is not the most common formulation of the problem but it fits the practical need described above\footnote{
  A more common formulation would be to maximize the value, with a penalty for exceeding some budget:
  $$\max_{a} U(a) = \max_{a} V_{\mathcal{I}}\left(a\right) - L\min\left(0, S_{\mathcal{I}}\left(a\right)-B\right).$$
This is equivalent in the sense that very similar first order conditions are derived from both approaches.}.
The \emph{Karush-Kuhn-Tucker} (KKT) conditions give the following\footnote{
KKT covers the case when the goal is attained, the more general result would be derived from first order sub-gradient condition for convex minimization.}:
\begin{align*}
a(x) f\left(a(x); x\right) &= \lambda v(x) f\left(a(x); x\right),&\quad \forall x,\\
a(x) &= \lambda v(x),&\quad \forall x,
\end{align*}
with $\lambda \in \left[0,K\right]$.

This means that the optimal strategy is to bid a value proportional to $v(x)$.
If we restrict to the case where the bidder tries to buy a certain amount of impression at the best possible price,
then the optimal strategy is to bid a constant bid. For the rest of this work, and without loss of generality,
the bidder aims at buying a certain amount of impression, hence all $u_i$ are $1$, $v(x)=1$ and
\begin{equation*}
a(x) = \lambda\in \left[0,K\right],\quad \forall x.
\end{equation*}

Most of these results do not hold for longer than $\delta$ if the auctions are no longer IID or $\iota_t$ non predictable.
In practice, random external factors affect the total volume of impressions $\iota_t$ in unpredictable ways.
The market conditions is also prone to large shifts, \eg{} when a very large campaign crowds out the other campaigns on a particular inventory.

As much as it is safe to assume everything is usually well predictable for the next few minutes, it is no longer the case for longer timescales.

\subsection{The bidding problem over a full day}

\begin{figure}[h]
\centering
\includegraphics[width=0.5\columnwidth]{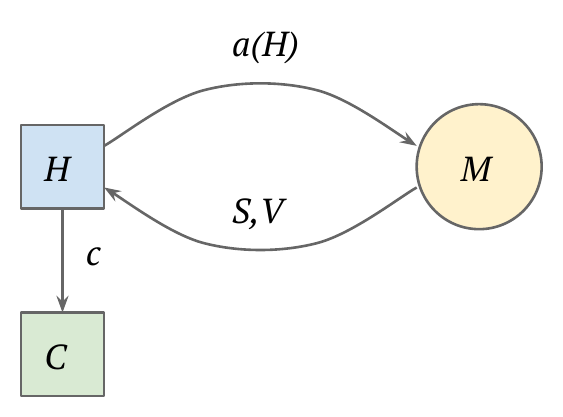}
\caption{In the MDP approach all the information available, e.g. all past spends $S$ and volume purchased $V$,
is embedded into the state $H$, the bid $a(H)$ is built out of $H$ and submitted to the market $M$.
Market response ($S$ and $V$) is fed back to the bidder and constitutes the reward/cost $c$,
adding up to the total cost $C$.}
\Description{In the MDP approach all the information available, e.g. all past spends $S$ and volume purchased $V$,
is embedded into the state $H$, the bid $a(H)$ is built out of $H$ and submitted to the market $M$.
Market response ($S$ and $V$) is fed back to the bidder and constitutes the reward/cost $c$,
adding up to the total cost $C$.}
\label{fig:mdp}
\end{figure}

Because the market is unpredictable, the bidder knows that no matter what he plans based on $H_t \in \mathcal{H}_t$,
he has to adjust constantly to new available information to reach its daily goal.
For this reason, the bidding strategy is now modeled as a \emph{Markov Decision Process} (MDP) as in \cite{gallego1994optimal, Karlsson2014, Karlsson2016, Karlsson2018, wu2018budget,cai2017real}, see Figure~\ref{fig:mdp}.

The full day is split in $T$ periods of duration $\delta$.
For each period, the bidder sets $a$ based on $t$, the remaining goal $G_t$
and a state $H_t$ coding for everything he learnt from past experience that is relevant,
and he observes the common distribution of all $b_i$ whose PDF and CDF are noted $f(b; H_t)$ and $F(b; H_t)$.
Note that the bid distribution is fully determined by the current state $H_t$.

The expected spend given knowledge of the state $H$ is
\begin{equation*}
S\left(H, a\right) = I\left(H\right) \cdot \int_0^{a}b f\left(b; H\right)\dif b
\end{equation*}
and the expected volume of impression is:
\begin{equation*}
V\left(H, a\right) = I\left(H\right) \cdot \int_0^{a} f\left(b; H\right)\dif b = I\left(H\right) \cdot F\left(a; H\right).
\end{equation*}

Let $C_t(G,H)$ be the lowest cost achievable to deliver $G$ impressions over $\left[t, T\right]$ given knowledge $H$.
\begin{equation*}C_t(G,H) = \min_{\left(a_s\right)_{s\geq t}} \mathbb{E}_t\left[ \sum_{s\geq t} S\left(H, a_s\right) + K\max\left(0, G-\sum_{s\geq t}V\left(H,a_s\right)\right)\right]
\end{equation*}
$C_t(G,H)$ is simply the sum of all spends $S$ plus the penalty
$K$ times the \emph{shortfall} given the optimal bidding strategy.

The optimal control is therefore fully characterized by the following Bellman equation:
\begin{equation}
  \begin{cases}
      C_t(G,H) = \min_a \mathbb{E}_t\left[S\left(H,a\right)
      + C_{t+1}\left(G-V\left(H,a\right), \mathcal{T}_t\left(H\right)\right)\right],\\
      C_T(G,H) = K\max\left(0, G\right).
  \end{cases}
  \label{eq:optimal_control}
\end{equation}
where $\mathcal{T}_t$ is the transition function, taking the current state $H_t$ and returning
the next state $H_{t+1} = \mathcal{T}_t(H_t, a_t)$.
Because $a$ has no impact on the market, the transition function can be noted $H_{t+1} = \mathcal{T}_t(H_t)$.

The first order condition on $a$ gives:
\begin{equation*}
  a_t\left(G,H\right) = \mathbb{E}_t\left[\frac{\partial}{\partial G}C_{t+1}\left(G-V\left(H,a_t\left(G,H\right)\right), \mathcal{T}_t\left(H\right)\right)\right].
\end{equation*}

It can be noted that at each period $t$, the bidder optimizes for the current period, knowing it has to optimize for the remaining periods, up to $T$. Also, the optimal $a$ is chosen equal to the marginal expected cost.
When the goal is far from being reached, $a$ will be high, else it will be low.

It can also be proved that $a$ is in the interval $\left[0,K\right]$. It is clearly the case for $t=T$,
but it is also the case for $t$ as long at is it true for $t+1$, because $C_t$ as a function of $G$ is a
mixture of $C_{t+1}$, which suffices to show
\begin{equation}
a_t \in \left[0,K\right],\quad \forall t.
\label{eq:bid_below_penalty}
\end{equation}

In the special case where $\delta$ is small enough and $H_t$ continuous, the problem can be usefully expressed in continuous time.

\subsection{Solution in continuous time}

Let us solve the optimal bidder problem in a continuous time setting with a simple Brownian motion model of available volume:
\begin{equation}
\dif H_t = \mu(t,H_t)\dif t + \sigma(t,H_t)\dif W_t,
\label{eq:volume_random_walk}
\end{equation}
where $W_t$ is a one-dimensional Wiener process.
The spend intensity at $t$ is
\begin{equation*}
S\left(H_t, a\right) = I\left(H_t\right) \cdot \int_0^{a}b f\left(b\right)\dif b
\end{equation*}
and the volume intensity is
\begin{equation*}
V\left(H_t, a\right) = I\left(H_t\right) \cdot \int_0^{a} f\left(b\right)\dif b = I\left(H_t\right) \cdot F\left(a\right).
\end{equation*}
In this model, the available volume is stochastic but the bid distribution is constant.

The minimization problem writes
\begin{multline*}
    C(t,G,H) = \min_{\left(a_s\right)_{t \leq s \leq T}} \mathbb{E}_t\left[ \int_t^T S\left(H_s, a_s\right)\dif s\right. \\
    \left.+K\max\left(0, G-\int_t^TV\left(H_s,a_s\right)\dif s\right)\right].
\end{multline*}
The \emph{Hamilton-Jacobi-Bellman} (HJB) equation states that
\begin{multline*}
    \frac{\partial C}{\partial t} + \frac{\partial C}{\partial H}\mu(t,H_t) + \frac{1}{2}\frac{\partial^2 C}{\partial H^2}\sigma^2(t,H_t)\\
    + \min_{a_t} \left[S\left(H_t, a_t\right) - \frac{\partial C}{\partial G}V\left(H_t, a_t\right)\right] = 0,
\end{multline*}
with the limit condition $C(T,G,H) = K\max\left(0, G\right)$.

At the minimum $a_t$ verifies the first order condition
\begin{equation*}
I\left(H_t\right) \cdot a_t f\left(a_t\right) =
\frac{\partial C}{\partial G}\left(t,G_t,H_t\right)I\left(H_t\right) \cdot f\left(a_t\right),
\end{equation*}
which reduces to:
\begin{equation*}
a_t = \frac{\partial C}{\partial G}\left(t,G_t,H_t\right).
\end{equation*}
The HJB equation can be solved in $a\left(t,G,H\right)$:
\begin{equation*}
  \frac{\partial a}{\partial t} + \frac{\partial a}{\partial H}\mu(t,H) + \frac{1}{2}\frac{\partial^2 a}{\partial H^2}\sigma^2(t,H)
   - I\left(H_t\right) \cdot F\left(a\right)\frac{\partial a}{\partial G} = 0,
\end{equation*}
with the limit conditions:
\begin{equation*}
\begin{cases}
a(T,G,H) = 0, & \text{if } G<0,\\
a(T,G,H) \in \left[0, K\right], & \text{if } G=0,\\
a(T,G,H) = K, & \text{if } G>0.
\end{cases}
\end{equation*}

\section{A numerical resolution}
\label{section:toy_problem}

In the special case where $\mu(t,H)=0$, $\sigma(t,H)=\sigma$,
$I\left(H\right)=H$ and $F$ is the CDF of a gaussian distribution, $a$ solves
the partial differential equation
\begin{equation}
  \frac{\partial a}{\partial t} + \frac{1}{2}\sigma^2\frac{\partial^2 a}{\partial H^2}
  - H \cdot F\left(a\right)\frac{\partial a}{\partial G} = 0,
  \label{eq:pde-sol}
\end{equation}
which can be solved numerically for various levels of uncertainty.

\begin{figure}[h]
\centering
\includegraphics[width=0.6\columnwidth]{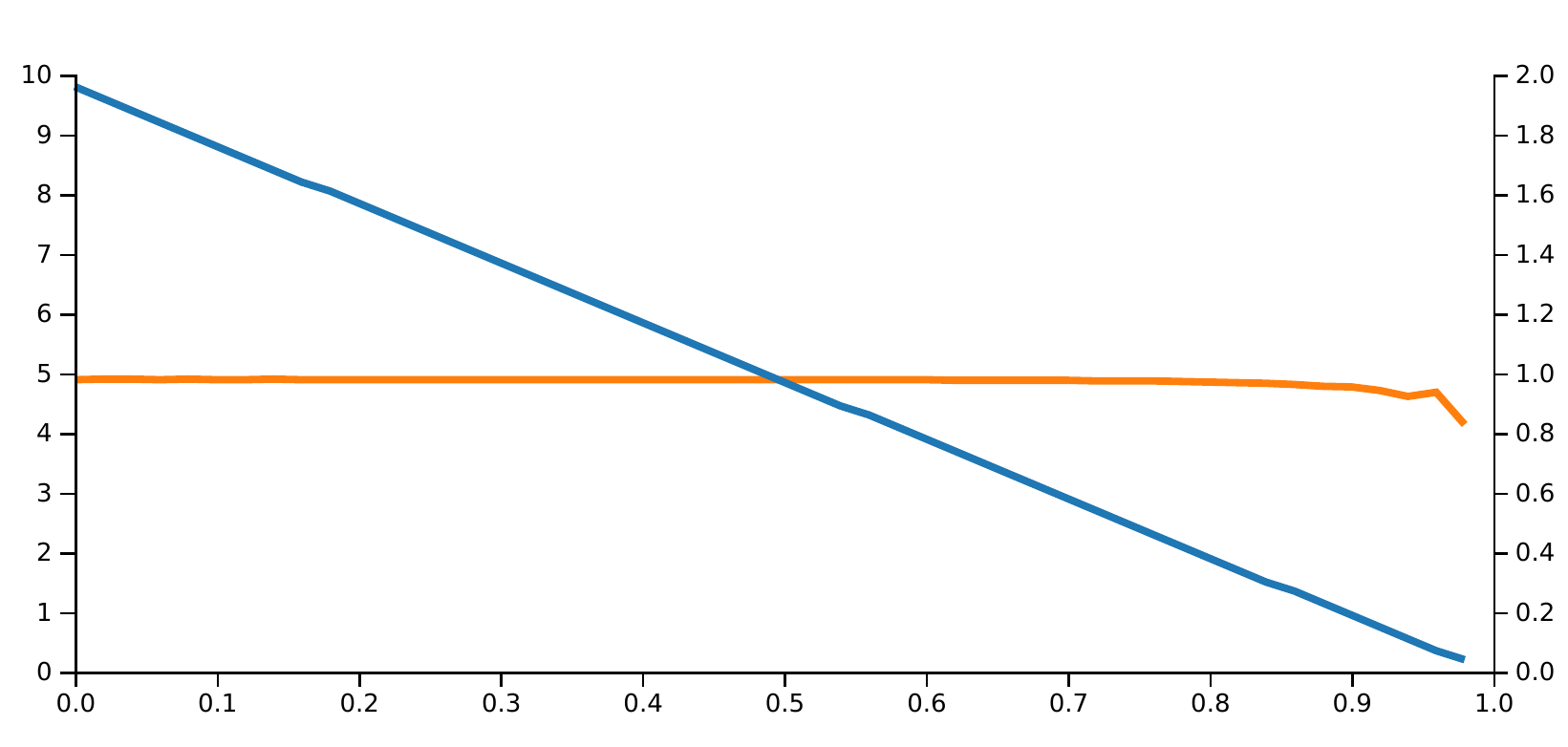}
\par$H$ constant accross the day\par\medskip
\includegraphics[width=0.6\columnwidth]{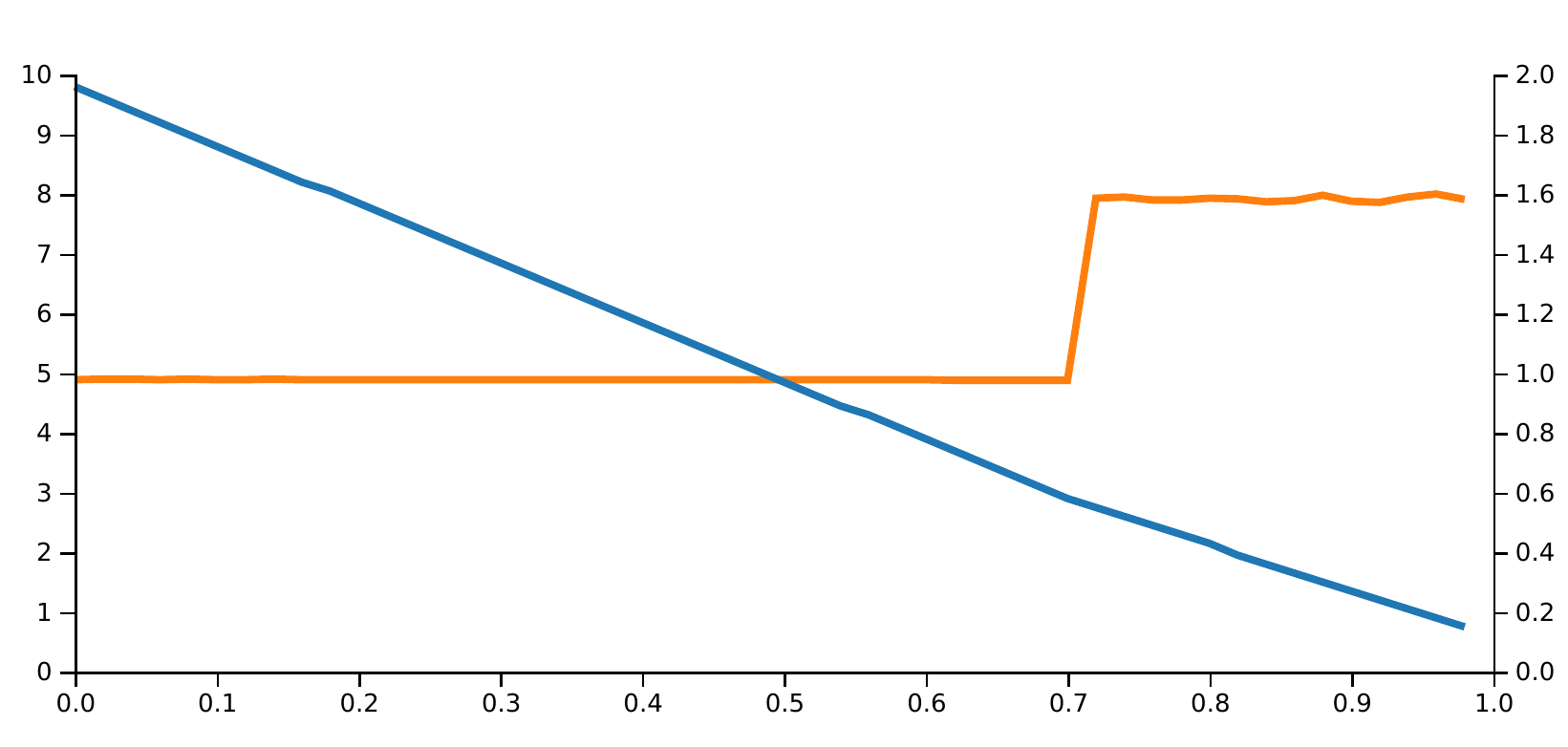}
\par$H$ with a 40\% drop at the end of the day\par\medskip
\caption{The resolution of Eq.~\ref{eq:pde-sol} with $\sigma=0$.
The blue curve shows the remaining goal to achieve across time. The orange curve shows the bid level.
One can notice the sharp increase in bid after the shock and the goal shortfall.}
\Description{The resolution of Eq.~\ref{eq:pde-sol} with $\sigma=0$.
The blue curve shows the remaining goal to achieve across time. The orange curve shows the bid level.
One can notice the sharp increase in bid after the shock and the goal shortfall.}
\label{fig:sigma-0-G-2}
\end{figure}

\begin{figure}[h]
\centering
\includegraphics[width=0.6\columnwidth]{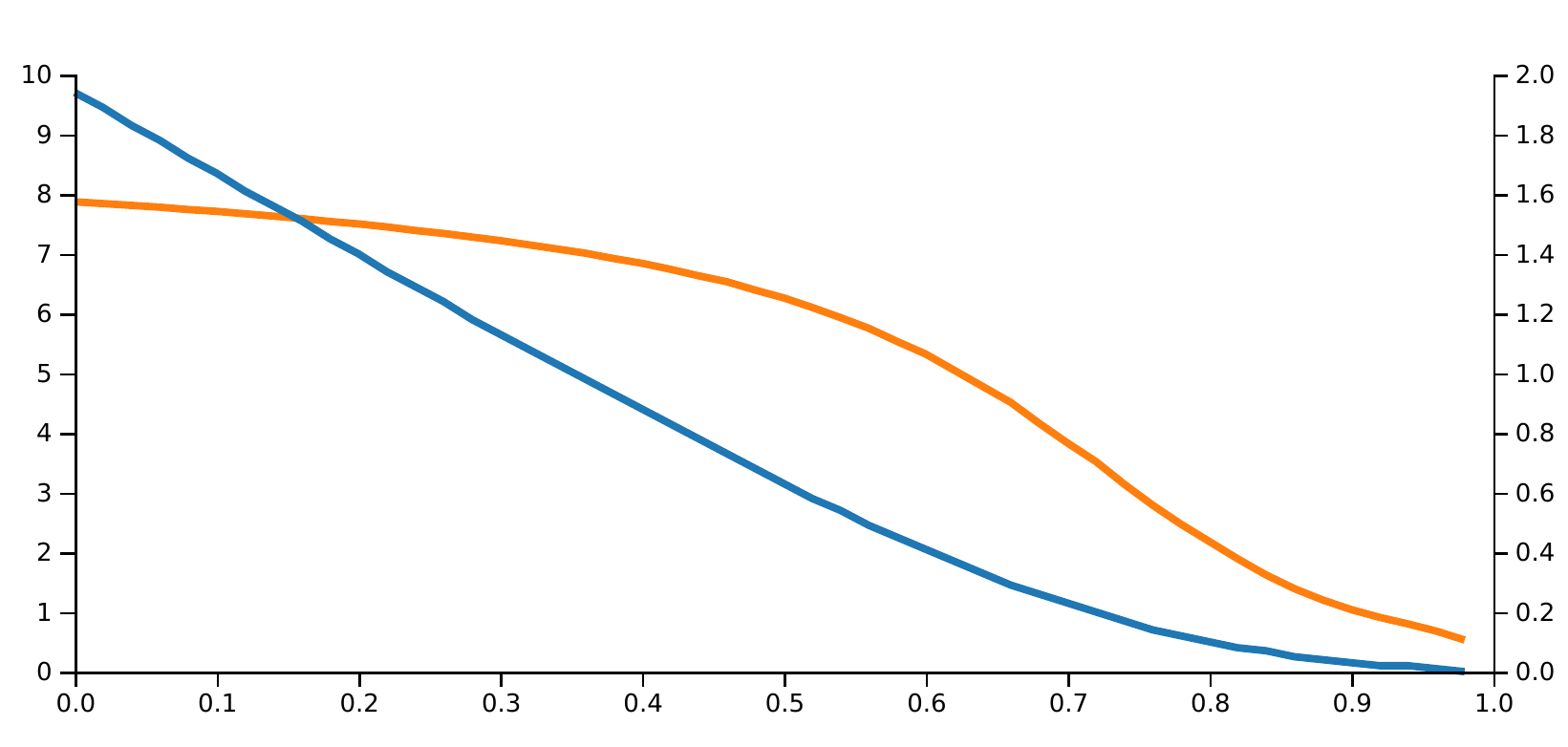}
\par$H$ constant accross the day\par\medskip
\includegraphics[width=0.6\columnwidth]{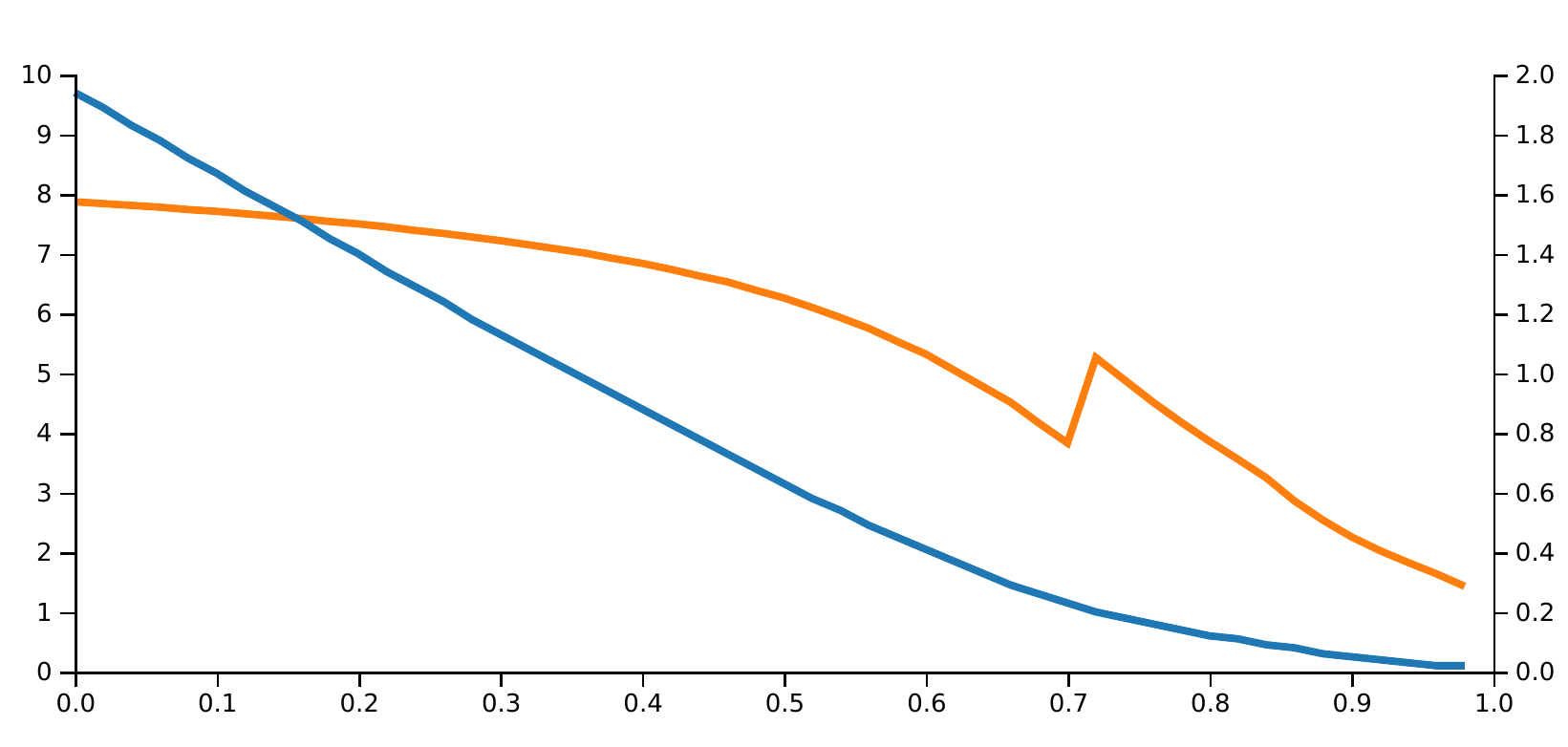}
\par$H$ with a 40\% drop at the end of the day\par\medskip
\caption{The resolution of Eq.~\ref{eq:pde-sol} with $\sigma=3$.
The bidder front-loads some of its delivery.}
\Description{The resolution of Eq.~\ref{eq:pde-sol} with $\sigma=3$.
The bidder front-loads some of its delivery.}
\label{fig:sigma-3-G-2}
\end{figure}

The numerical solutions in Fig.~\ref{fig:sigma-0-G-2} and Fig.~\ref{fig:sigma-3-G-2} show that the
introduction of uncertainty $\sigma=3$ in the market induces some provisioning behavior in the optimal strategy.
This provisioning for risk is materialized by a decreasing bid with time whenever the risk does not materialize,
which is the case in those simulations where $H$ is constant except for a shock at the end of the day.

The MDP approach solves the bidding problem under uncertainty in a satisfactory way, but in
the general case the relevant information in all the information available: $H$, is not obviously observable and
using all $H$ requires working in spaces too large to be practical.

The general case can be approximated. Such an approximation needs to be chosen in a functional space
rich enough to capture the desired features.
In Section~\ref{section:mdp}, RNNs are tested to this end.

\section{Practical MDP bidding}
\label{section:mdp}

As demonstrated above, a robust bidding strategy should adjust its bidding behavior continuously based on the last
information available, but also based on the fact it has to adjust in the future.

Building such a system is complex. Let us say a bidder records the history of all the bids $b_i$, spends $S_i$, purchased volumes $V_i$ and any other relevant information $x_i$ for each auction $i$.
This sequence is noted
\begin{equation*}
H_i = \left(X_0, X_1, \cdots, X_{i-1}, X_i\right) \in \mathcal{X}^*,
\end{equation*}
where $X_i = \left(b_i, S_i, V_i, x_i\right) \in \mathcal{X}$.

A bidding strategy should be a function $a\left(t,G,H\right)$ of time, remaining goal and the finite sequences $H\in \mathcal{X}^*$ to $\mathbb{R}$.
Solving a minimization problem on such a space is largely intractable, even numerically,
so we rely on some finite dimensional representation $\FS \in \mathbb{R}^n$ of $H$, enabling a fair approximation of the solution:
\begin{equation*}
a\left(t,G,\FS\right) \approx a\left(t,G,H\right).
\end{equation*}

\begin{figure}[h]
\centering
\includegraphics[width=0.5\columnwidth]{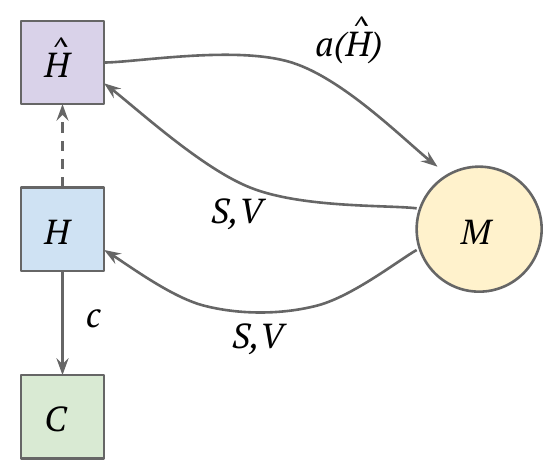}
\caption{In the practical MDP approach all the information available, e.g. all past spends $S$ and volume purchased $V$ are used to update a finite dimensional \emph{proxy} state $\FS$.
The bid $a(\FS)$ is submitted to the market $M$.
The state transition function (the way $\FS$ is updated) is trained from data to minimize total cost $C$.}
\Description{In the practical MDP approach all the information available, e.g. all past spends $S$ and volume purchased $V$ are used to update a finite dimensional \emph{proxy} state $\FS$.
The bid $a(\FS)$ is submitted to the market $M$.
The state transition function (the way $\FS$ is updated) is trained from data to minimize total cost $C$.}
\label{fig:mdp-proxy}
\end{figure}

The state $\FS_t$ is not updated for every auction, but instead at a regular pace.
It is computed based on $\FS_{t-1}$, the remaining time to deliver $T-t$,
the remaining volume to reach $G_t = G_{t-1} - V_{t-1}$ and the last spend $S_{t-1}$:
\begin{equation*}
\FS_t = \mathcal{T}\left(\FS_{t-1}, V_{t-1}, S_{t-1}; \theta\right).
\end{equation*}
The transition function $\mathcal{T}(\cdot;\theta)$ is trained to minimize the cost of the campaign (\cf{} Fig.~\ref{fig:mdp-proxy}).

In the next two sections, two different practical implementations of a bid controller to provide an approximation of the solution are presented:
the \emph{Proportional Integral controller} in Section~\ref{section:pi}, and the \emph{Recurrent Neural Network controller} in Section~\ref{section:rnn}.

\subsection{The Proportional Integral controller}
\label{section:pi}

The Proportional Integral (PI) controller\footnote{See \cite[Chapter~10]{Astrom2008} for an introduction to the PI control.} is widely used in various industries~\cite{Desborough2002}. \cite{Karlsson2014,Zhang2016} propose to apply it to the bidding problem.

The interaction between the bidding agent and the market can be modeled as a feedback system composed of a \emph{feedback controller} and a \emph{block} $\market$ representing the RTB market.
The system receives as input a \emph{reference signal} $\rsignal_t = \rsignal\left(t, \goal_t; \theta_V\right)$: a target volume for the next time period.

From the feedback  $\volume_t$ received from the market, the controller computes a \emph{control error} $e_t=\rsignal_t-\volume_t$. Based on it, the controller maintains a state
and uses it to generate a new control variable (or action of bidding at a specific bid level) $\cv_t$
\begin{equation}
  \cv_t = \theta_P e_t + \theta_I \sum_{s=0}^{t}e_s, 
  \label{eq:pid}
\end{equation}
where $\theta_P$, $\theta_I$ are two parameters called the \emph{proportional} and \emph{integral gains}.

In the PI setup, the state and its transition function $\mathcal{T}(\cdot;\theta)$, where $\theta = (\theta_{V}, \theta_P, \theta_I)$, can be expressed
\begin{equation}
\FS_t = \begin{pmatrix} 0, 0, 0\\ 0, 1, 0\\ 0, \theta_I, 0\end{pmatrix} \FS_{t-1}
+ \begin{pmatrix} 1\\ 1\\ \theta_P + \theta_I\end{pmatrix} \left( \rsignal\left(t-1, \goal_{t-1}; \theta_V \right)- \volume_{t-1} \right)
\end{equation}
and
\begin{equation}
a(t, \goal_t, \FS_t) = \langle \begin{pmatrix}0\\0\\1\end{pmatrix}, \FS_{t}\rangle.
\end{equation}

Training the PI controller is done in two steps:
a reference forecasted volume process $\rsignal$ is defined and trained,
and then the gains are tuned using Stochastic Gradient Descent.

Although simple and robust, this approach comes with some flaws.
\begin{enumerate}
  \item It depends on a separate forecaster $\rsignal$.
  \item It is designed to target a current value, not to optimize a lifetime cost function;
  this is mitigated by the fact the parameters are tuned against the lifetime cost.
  \item The uncertainty about the market is not modeled, it is barely taken into account through
  the cost function, but no component of the state $\FS$ really reflects anything about risk.
  \item The important gap between the small number of parameters of the PI model
  and the large amount of data available suggests probable underfitting.
  Capacity can be added to the model by allowing adaptive gains, setting thresholds and special cases,
  but those are merely local patches.
\end{enumerate}
To overcome these flaws, we introduce in the next section a new approach leveraging a Recurrent Neural Network to approximate the bidding problem solution. A PI controller is used as benchmark to the RNN approach.

\subsection{The RNN controller}
\label{section:rnn}

The Recurrent Neural Network (RNN)\footnote{See \cite[Chapter~10]{Goodfellow-et-al-2016}.} controller unit used in all the experiments presented in this paper is a Gated Recurrent Unit (GRU, see~\cite{cho2014learning})\footnote{Long Short-Term Memory (LSTM, cf~\cite{hochreiter1997long}) was also experimented with but the results were similar to the GRU ones.},  with
\begin{description}
\item[input:] a vector$\left(T-t, \goal_t, \volume_t, \spend_t\right)$,
\item[state:]  a vector $\FS$ with dimension 16\footnote{Simple trials were also conducted to assess the interest of using more neurons in recurrent units in the RNN architecture, be it wider or deeper. No significant gain was found, and a detailed assessment lies beyond the scope of this paper.},
\item[activation:] a hyperbolic tangent function rescaled for the first component of the state $\FS$ to be between $0$ and the penalty level $\pen$\footnote{$\pen$ is the highest possible bid in an optimal strategy, cf Eq.~(\ref{eq:bid_below_penalty})},
\end{description}
and where the bid level is given by the first component of the state $\FS$ of the GRU layer:
\begin{equation}
a(t, \goal_t, \FS_t) = \langle\begin{pmatrix}1\\0\\ \vdots \\0  \end{pmatrix}, \FS_{t}\rangle.
\end{equation}

Through the recurrent connections the model can retain information about the past in its state, enabling it to discover temporal correlations between events and its interactions with the environment even when these are far away from each other in the data. Using a RNN allows to take advantage of a much richer family of functions to learn an approximate solution to the bidding problem.

\section{Experiments}
\label{section:experiments}

\subsection{Practical setting}

In practice, the massive number of auctions occurring simultaneously makes unrealistic
the resolution of the optimal control~\eqref{eq:optimal_control} for each auction and campaign. Fortunately, taking periodic control decisions (\eg{} every 5 minutes) on aggregated feedback is sufficient. It is thus possible to handle a very large number of campaigns, with the following steps:
\begin{itemize}
  \item at the beginning of each period, choose a level for the control variable $\cv$,
  \item get an aggregated feedback (realized volume and spend) from the previous period in response to the level $\cv$.
\end{itemize}

This kind of architecture introduces discontinuity in the response of the controlled advertising system.
Yet, the problem can be turned into a continuous control problem (\cf~\cite{Karlsson2016,Karlsson2018}).

\begin{figure}[h]
\centering
\includegraphics[width=\columnwidth]{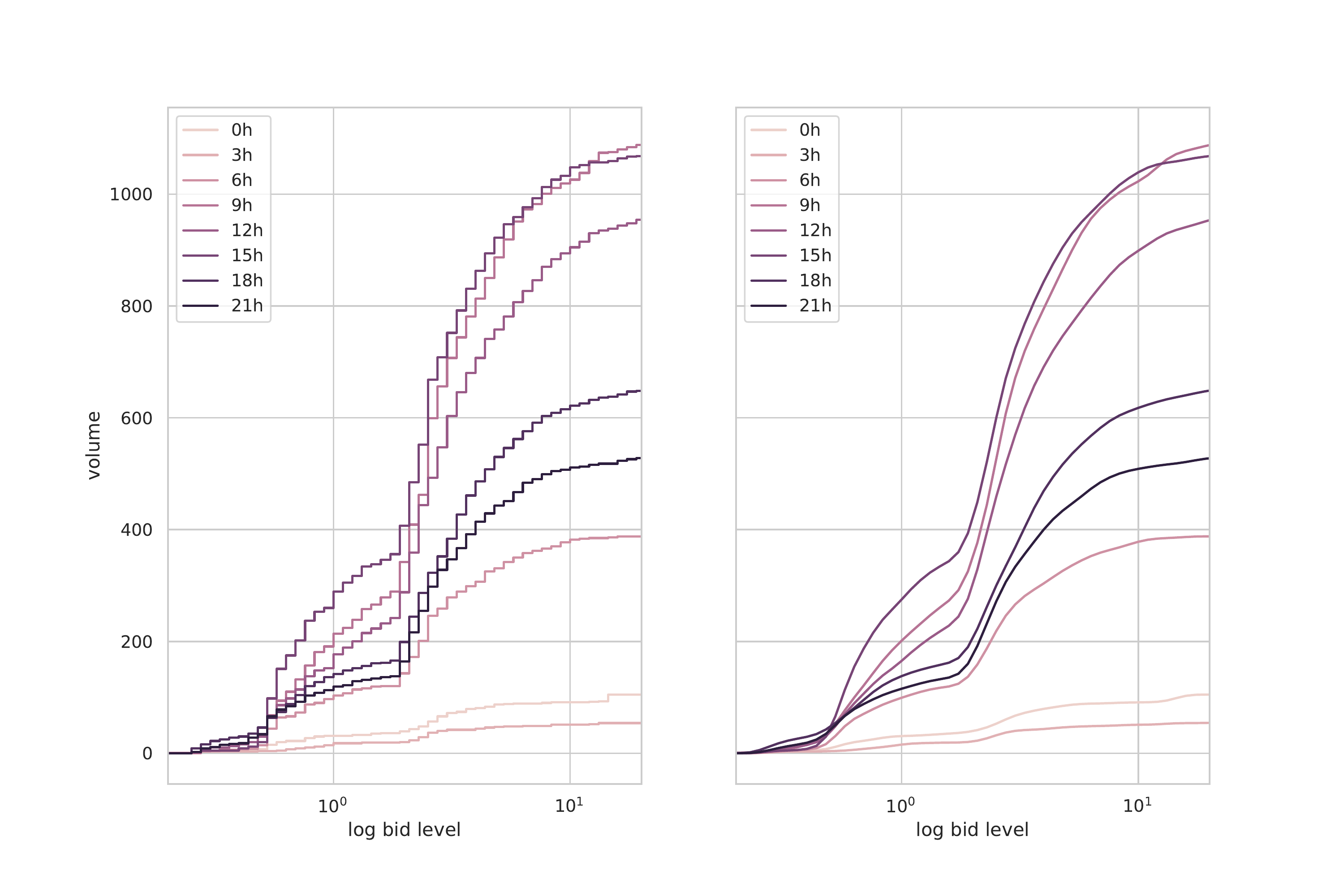}
\caption{An example of bid level to volume mapping evolution through the day. Left: no noise (Dirac bids). Right: volume obtained by Gamma-distributed bids around each bid level.}
\Description{Left plot shows increasing step functions mapping a bid level to the available volume on the market. For a given bid level, the volume varies during the day. Right plot shows the same price-volume relationship smoothed by the convolution with the bid distribution at each price point.  }
\label{fig:bid-volume-mapping}
\end{figure}

Furthermore, response curves exhibit discontinuities (\cf~Fig.~\ref{fig:bid-volume-mapping}, left plot).
These discontinuities can be smoothed out (right part of Fig.~\ref{fig:bid-volume-mapping}) by not bidding a constant bid level during the time period but by drawing bid prices according to some distribution  (\eg{} log-normal, Gamma, \etc{}) around the control variable $\cv$,
such as proposed in \cite{Karlsson2014, Karlsson2018}.

This leads to the following loop:
\begin{itemize}
  \item at the beginning of each period, choose a level for the control variable $\cv$,
  \item for each auction occurring in the period, draw a bid price according a distribution based on $\cv$.
  \item get an aggregated feedback (realized volume and spend) from the previous period.
\end{itemize}

Note that contrary to the general setup of the bidding problem that suffers from censorship due to the ad auction selection (\cf~\cite{Zhang2016}),
one can alleviate this particular issue in this practical setup since publisher data is available.
In the absence of uncensored data, the bid randomization would further help, by realizing some of the
exploration effort in the \emph{explore vs exploit} dilemma introduced by bid-dependent censorship.


\subsection{Data used}
Two types of datasets are used in the numerical experiments presented in this paper:
\begin{enumerate}
  \item Simulated synthetic data sets, generated by applying various transformations (shocks or random walk) throughout time to a base linear volume curve, that helps in appreciating salient features of the RNN models,
  \item Production RTB data, constituted by 5-minute snapshots of the actual price-volume mapping
  for all display ad placements with significant daily volume\footnote{Restricted to the placements with a minimum of 1000 daily impressions.}
  from large publishers on one of the leading \emph{Supply Side Platform} (SSP) and Ad Exchange globally.
  The ad placements here can be seen as proxys of segments targeted by campaigns. In production, the RNN would have to be trained on currently running campaigns.
\end{enumerate}

The production dataset is created using logs from actual RTB auctions run on around 1000 ad placements over 8 days, containing about 115M won impressions. All these impressions are used to build winning bid distributions for 5-minutes periods over a full day ($T=288$) of each ad placement. The winning bid distributions are discretized on a CPM bid scale with 100 exponential increments between $0.01$ and $100$.

For offline training and evaluation purposes on production data, a bidding problem instance is comprised of a random draw of an actual bid-volume mapping process and
of a random volume goal, uniformly drawn between 10 and 1000. The controller therefore is exposed to scenarios with not enough of volume to meet the target given the penalty level, as well as scenarios where enough volume was available.

The production data is split into non-overlapping training, validation and evaluation datasets using different days. The training set of models on production data contains 1 million different bidding problems and evaluation is performed on a set of around 110K bidding problems of increasing difficulty.
For the simulated case study, given the simplified setting, training is stopped after learning from 20K bidding problems.

\subsection{Training and evaluation}

The implementation of both the benchmark (PI controller) and the RNN controller is done in TensorFlow~\cite{tensorflow2015-whitepaper}.
The input data instances are randomly shuffled and processed by batches of size 100.

The aim is to minimize the total cost of a campaign, so the training loss $\loss$ is composed of the sum of the spend and the penalty terms over a full day:
\begin{equation}
  \loss := \TotalSpend{T} + \pen \max\left( 0, \goal - \TotalVolume{T} \right)
  \label{eq:loss}
\end{equation}
where the spend $\spend_t$ and volume won $\volume_t$ are computed from the bid level $a_t$ of the MDP bidding controller by simulating the feedback using the input bid distribution at each time step and propagating the state over the full sequence of time periods.

Models are trained using \emph{Stochastic Gradient Descent} (SGD) with an inverse-time decay of the learning rate\footnote{The learning rate schedule is the following: $\alpha(n) = \frac{\alpha_0}{1 + \eta \lfloor\frac{n}{N}\rfloor}$ with initial learning rate $\alpha_0 = 0.1$, decay rate $\eta = 0.5$, and decay steps $N = 400$.}.
To help alleviate possible exploding gradients issues, gradient clipping is used as described in~\cite{Pascanu2012}. Other more sophisticated optimization methods have been tried without significant impact on the results.

Cross-validation is performed regularly during the training on a fixed set of validation bidding problems and the optimal model parameters are picked as the best evaluation seen on the validation set during the training. In practice, no model presented any overfitting issue as performance results generalized well to unseen data.

\subsection{Numerical results}
\label{section:numerical_results}
\subsubsection{Simulated case study}

\begin{figure*}
  \centering
  \includegraphics[scale=0.55]{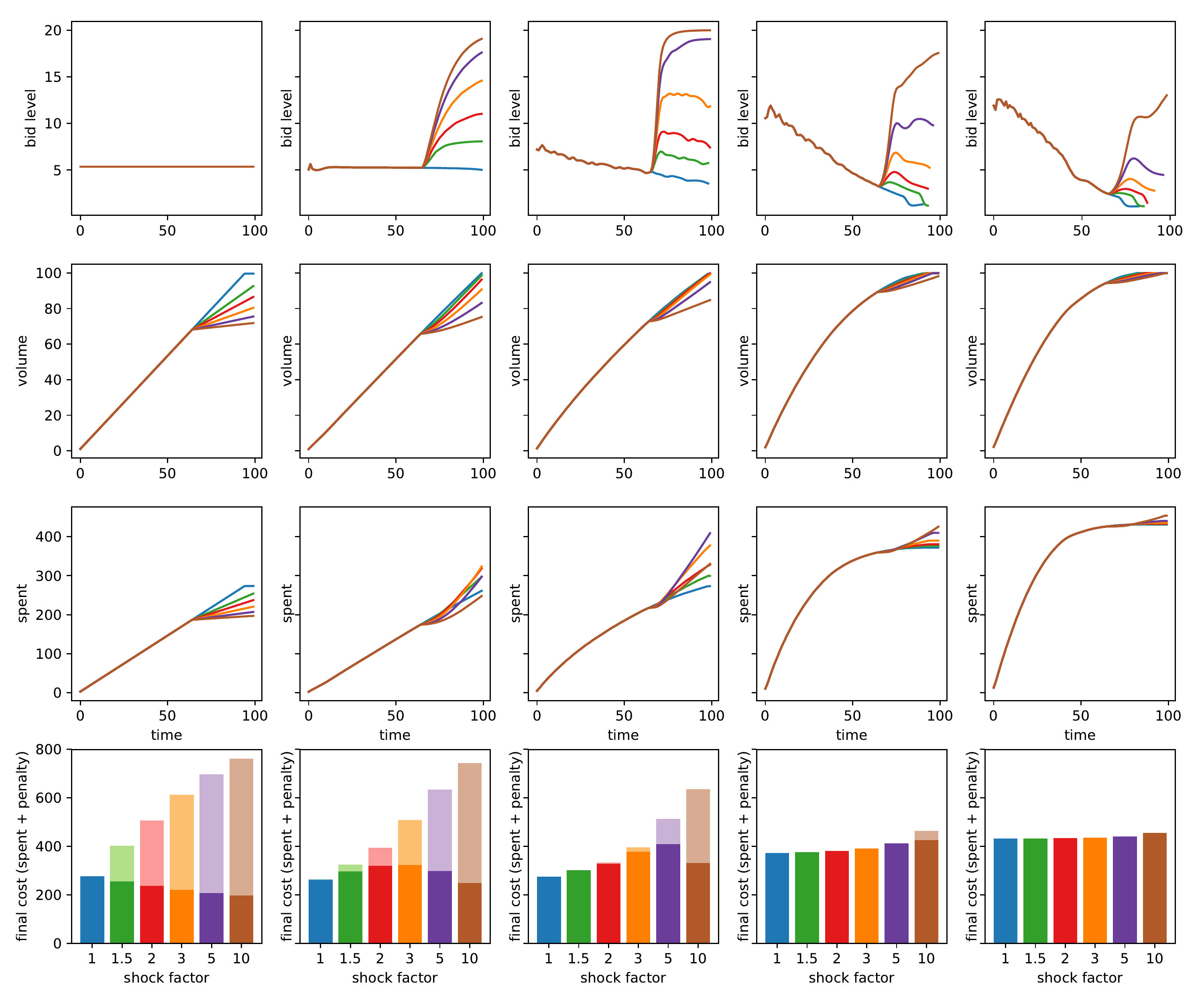}
  \Description{Bid strategy is constant when noise was absent during the training.
    Hence the volume acquired and spend linearly grow through time in the scenario without any shock in the available volume.
    In scenarios with permanent shock factors, this model carries on bidding at the same level, thus falling short of the volume target of $100$ impressions.
    As the final volume shortfall grows with the shock factor, the bigger the shock occurs, the bigger the end penalty is.
    It can be observed that the more noise a RNN controller model was trained with, the higher it starts bidding and the easier it absorbs shocks in the available volume.
    The models trained with more uncertainty in the available volume tend to provision for the shortfall risk by bidding higher and earlier than the optimal level in the absence of noise.
    To the extreme, the optimal solution is to buy as fast as one can.
    This obviously entails a higher spend when the risk does not materialize.
    However, the more risk was anticipated by a model through its training, the lower the final spend and penalty are when risk do realize, amplified here by increasing level of permanent shocks.
  }
  \caption{Bid strategy of RNN controller models trained with an increasing level of noise, and their reaction to permanent volume shock scenarios.
  Each column shows the bid level, volume and spend throughout time as well as the final cost (sum of spend and penalty) of the same RNN model trained on simulated data with an increasing level of noise.
Columns 1 to 5 respectively correspond to standard deviations of $0$, $0.2$, $1.$, $5.$, and $10.$ during the training.
The models are here evaluated on deterministic data without noise. In each column, the six lines or bars represent various shock factors impacting the available volumes for all dates $t \geq 65$.
A shock factor of $x$ means that after the shock the available volume given the same bid level is divided by $x$.
The bottom row decomposes the final cost incurred into the spend for buying the final impressions (bottom dark-colored bars)
and the penalty that may be received if the volume target is not reached (light-colored stacked bars).    }
  \label{fig:volume_shocks_increasing_noise}
\end{figure*}

A first experimental setup on simulated data goes back to the simplified setting from Section~\ref{section:toy_problem}.
Figure~\ref{fig:volume_shocks_increasing_noise} displays the bid level, volume and spend during the day, along with the final delivery cost decomposed into the final spend and penalty for five different RNN models.
Each RNN model is trained on a simulated dataset for which the volume process follows Eq.~\ref{eq:volume_random_walk} with a different level of noise (and no drift term).
Each column evaluates the same model on six cases without noise, for which a permanent shock in the available volume happens for all dates $t \geq 65$.
A shock factor of $\delta$ means that after the shock the available volume given the same bid level is divided by $\delta$.

Note that the RNN model is able to learn a good approximation of the optimal strategy derived theoretically.
The RNN controller exhibits the same behavior as the one evidenced in Figures~\ref{fig:sigma-0-G-2} and \ref{fig:sigma-3-G-2} by exact resolutions of the control problem.
Indeed, the bid strategy is constant when noise is absent during the training. Hence the volume acquired and spend linearly grow through time in the scenario without any shock in the available volume.
In scenarios with permanent shock factors, this model carries on bidding at the same level, thus falling short of the volume target of $100$ impressions.
As the final volume shortfall grows with the shock factor, the bigger the shock occurs, the bigger the end penalty is.

One can observe that the more noise a RNN controller model was trained with, the higher it starts bidding and the easier it absorbs shocks in the available volume.
The models trained with more uncertainty in the available volume tend to provision for the shortfall risk by bidding higher and earlier than the optimal level in the absence of noise.
To the extreme, the optimal solution is to buy as fast as one can (without bidding higher than the penalty).
This obviously entails a higher spend when the risk does not materialize.
However, the more risk was anticipated by a model through its training, the lower the final spend and penalty are when risk do realize,
demonstrated here by increasing level of permanent shocks.

In a second experimental setup on simulated data, two RNN models were trained with either a low or high\footnote{The high noise scenario is calibrated to be consistent with the average level of daily noise observed on our real data, its interquartile range being $[7., 18.]$. So the low noise scenario does represent a situation where risk is deeply underestimated.} level of noise and evaluated under both a low or high noise scenario.
Main results are shown in Table~\ref{table:high_low_noise}.
The experiment shows as expected that for both models the cost of delivery and the bidding problem difficulty (probability of shortfall) are higher in the evaluation scenario with high noise. However, given an evaluation scenario, i.e. an amount of noise that realizes, the delivery cost is lower for the model that has been trained on data with the same amount of uncertainty.

\begin{table}
\begin{tabular}{cccc}
\multicolumn{4}{c}{Mean final cost (std. dev.), shortfall probability} \\
\hline
& & \multicolumn{2}{c}{Evaluation scenario} \\
& & $\sigma=0.1$ & $\sigma=10.$ \\
Learning & $\sigma=0.1$ & \$271 (\$13), 5\% & \$915 (\$516), 82\% \\
scenario & $\sigma=10.$ & \$431 (\$9), 0\% & \$764 (\$405), 57\% \\
\end{tabular}
\caption{Simulation results of RNN models trained and evaluated under high or low noise scenarios.}
\label{table:high_low_noise}
\end{table}

\subsubsection{Training on production data}
Performance results on actual market data are available in Table~\ref{table:real_world_results}.
As a benchmark, a PI controller is tuned to follow a reference pacing curve fitted on the training data.
Indeed, a good approximation of the internet traffic intraday seasonality can be obtained using a model with only two harmonics \cite{Karlsson2014}.

Table~\ref{table:real_world_results} details the average total cost of delivering campaigns of increasing daily volume goals for both the PI model and the RNN model. Overall, the RNN model is able to reduce delivery cost by about 20\% compared to the PI model.
As the volume target increases, so does the bidding problem difficulty as the total available volume under the penalty level is constraining the bid strategy for a larger share of the dataset. Eventually, for very large volume goals the optimal strategy is to bid the penalty level, capturing all the volume below this level and paying the penalty for each missed impression. Thus lower performance improvements are expected for the larger goals, relative to the size of the targeting.

\begin{table}
\begin{tabular}{lrrr}
\toprule
{} &  \multicolumn{2}{c}{Delivery cost (\$CPM)} &  ratio \\
Goal (imps) &   PI  &  RNN & RNN/PI \\
\midrule
100           &                  1.02 &                   0.82 &                           0.80 \\
500           &                  1.28 &                   1.02 &                           0.79 \\
1000          &                  1.60 &                   1.27 &                           0.80 \\
1500          &                  2.06 &                   1.76 &                           0.85 \\
\bottomrule
\end{tabular}
\caption{Model performance comparison on actual market data.}
\label{table:real_world_results}
\end{table}

\section{Conclusion and future works}

The RNN controller model proposed in this paper provides both an effective and practical method to solve the optimal bidding problem. It has the advantage not to rely on manually engineered features to represent knowledge about the current state or history that could be leveraged in a bidding strategy, but instead infers it from the data.
For instance, a more advanced, adaptative, PI controller could be employed to tackle the control problem, \eg{} by using splines modeling the price-volume mapping to efficiently store the response gain at various price points.
However, such a model would still lack useful elements from the very complex state space it evolves in, mainly because it overlooks the impact that uncertainty about future market volume and bid landscape has on the optimal strategy.

Numerical experiments demonstrate that the proposed approach is able to improve significantly on existing bidding controllers, while being trainable and usable at production scale. The approximation of the state and space transition provided by the RNN leads to a solution that captures a key aspect of the solution, namely provisioning against the risk of underdelivery.

This work could be extended in many ways:
\begin{itemize}
  \item The observability of all bids including those of lost auctions is convenient in the case
  of our work. This assumption could nevertheless be relaxed. The reconstruction of bid distributions
  for training would probably be more complex and the noise added to the bid would need to be used as an exploration device.
  \item In practice, setting multiple goals would be an interesting feature to add,
  e.g. buying impressions with some guarantee of viewability. The equations would be marginally changed.
  \item The first price and exotic auction cases add a significant amount of complexity to this approach,
  however those questions would be resolved at the impression scale, while the macroscopic (5 min) scale control problem would probably hold in a similar way.
  \item Giving the RNN some more feedback, based on the noise injected in the bid could probably help.
\end{itemize}

More generally, this paper shows how RNNs can be applied to complex control problem with success.

\bibliographystyle{ACM-Reference-Format}
\bibliography{deep_learning_controller.bib}{}



\end{document}